\PassOptionsToPackage{table,dvipsnames}{xcolor}
\documentclass[10pt,twocolumn,letterpaper]{article}

\usepackage{iccv}              

\definecolor{iccvblue}{rgb}{0.21,0.49,0.74}
\definecolor{cellgray}{gray}{0.9}
\usepackage[pagebackref,breaklinks,colorlinks,allcolors=iccvblue]{hyperref}

\usepackage{multirow}
\usepackage{pifont}
\usepackage{cuted}
\usepackage{xcolor}
\usepackage{subcaption}
\usepackage[accsupp]{axessibility}

\newcommand{\ourmodel}{\textsc{RAS}\xspace}
\newcommand{\ourlargedata}{\textsc{MaskGroups-2M}\xspace}
\newcommand{\ourgooddata}{\textsc{MaskGroups-HQ}\xspace}

\newcommand{\cmark}{\ding{51}} 
\newcommand{\xmark}{\ding{55}} 

\title{Refer to Any Segmentation Mask Group With Vision-Language Prompts}

\author{%
Shengcao Cao$^1$ \quad Zijun Wei$^2$ \quad Jason Kuen$^2$ \quad Kangning Liu$^2$ \quad Lingzhi Zhang$^2$ \\
Jiuxiang Gu$^2$ \quad HyunJoon Jung$^2$ \quad Liang-Yan Gui$^1$\thanks{Equal advising.} \quad Yu-Xiong Wang$^{1*}$ \\
$^1$University of Illinois Urbana-Champaign \quad $^2$Adobe \\
{\small\texttt{\{cao44,lgui,yxw\}@illinois.edu\quad\{zwei,kuen,kangningl,lingzzha,jigu,hjung\}@adobe.com}}
}

\begin{document}

\maketitle

\begin{strip}
    \centering
    \vspace*{-12mm}
    \includegraphics[width=\linewidth]{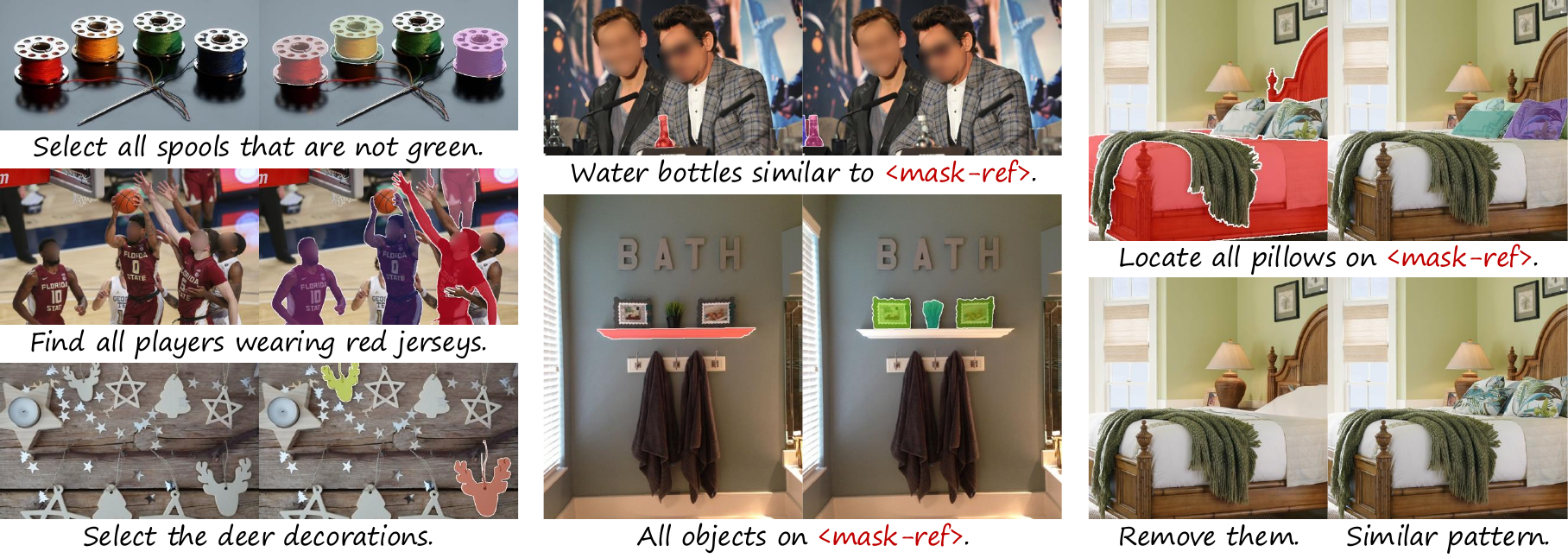}
    \captionof{figure}{\textbf{Omnimodal referring expression segmentation (ORES) according to arbitrary vision-language prompts.} \textit{(Left)} Our approach, Refer to Any Segmentation Mask Group (\ourmodel), can understand a complex text prompt involving multiple conditions. \textit{(Middle)} Reference visual entities can be included as visual prompts to enhance expressivity, addressing the challenge of describing the same details using language alone. \textit{(Right)} The grouped segmentation masks conveniently enable various fine-grained downstream applications, such as object removal and editing. In each pair of images, the left one is the input and the right one is the output. Best viewed on an electronic device with zoom-in functionality.}
    \label{fig:teaser}
\end{strip}

\begin{abstract}
Recent image segmentation models have advanced to segment images into high-quality masks for visual entities\footnote{Countable objects and amorphous stuff regions~\cite{kirillov2019panoptic, qi2022open}.}, and yet they cannot provide comprehensive semantic understanding for complex queries based on both language and vision. This limitation reduces their effectiveness in applications that require user-friendly interactions driven by vision-language prompts.
To bridge this gap, we introduce a novel task of omnimodal referring expression segmentation (ORES). In this task, a model produces a group of masks based on arbitrary prompts specified by text only or text plus reference visual entities. To address this new challenge, we propose a novel framework to ``Refer to Any Segmentation Mask Group'' (RAS), which augments segmentation models with complex multimodal interactions and comprehension via a mask-centric large multimodal model.
For training and benchmarking ORES models, we create datasets \ourlargedata and \ourgooddata to include diverse mask groups specified by text and reference entities. Through extensive evaluation, we demonstrate superior performance of RAS on our new ORES task, as well as classic referring expression segmentation (RES) and generalized referring expression segmentation (GRES) tasks.
Project page: \url{https://Ref2Any.github.io}.
\end{abstract}
\vspace{-2mm}

\section{Introduction}
\label{sec:intro}

Referring expression segmentation (RES)~\cite{yu2016modeling, liu2023gres, hu2016segmentation} enables language-based object segmentation by relating a text prompt with a segmentation mask for the referred target, and can be generalized (\ie, GRES~\cite{liu2023gres}) for multiple targets.
However, real-world applications, such as autonomous driving~\cite{shao2024lmdrive,cui2024survey}, robotics~\cite{gao2024physically,liu2024vision, xu2024survey}, augmented reality~\cite{konenkov2024vr}, and image editing~\cite{shen2024empowering, nguyen2023visual,guo2024prompthis}, require additional flexibility in the prompt. As exemplified in Figure~\ref{fig:teaser}, a user may want to specify a segmentation instruction that involves a \emph{relationship/comparison/interaction with a reference visual entity} (see red regions in the middle column). Expressing the reference via visual and text prompts together is usually preferred over text-only descriptions in such cases, because language may not be able to concisely and accurately locate the reference entity or describe its intricate characteristics in a complex image.

In this work, we introduce a novel task of \textbf{omnimodal referring expression segmentation (ORES) with vision-language prompts} (Figure~\ref{fig:teaser}). For a given image, the system generates a \emph{group} of relevant masks that satisfy a user-specified instruction, which can be
a) a text-only prompt describing a property (\eg, category, attribute, position, or their combination) of the targets, or
b) a vision-enriched prompt that provides masks of reference entities and expresses a complex property involving the reference entities. 

Unfortunately, \emph{it is not straightforward to extend existing models for this new challenge} (Table~\ref{tab:comparison}).
Interactive segmentation models (\eg, SEEM~\cite{zou2024segment}) accept both text and visual prompts, but visual prompts only lead to the directly indicated entity rather than any related ones. In contrast, ORES aims to return a group of relevant masks based on the prompt, providing a more contextually cohesive response to the user's input. The visual prompt can be provided to conveniently describe a relationship between the target(s) and a reference visual entity that is difficult to describe verbally.
Some grounding large multimodal models (LMMs) accept mask or region prompts~\cite{you2024ferret, rasheed2024glamm, zhang2024groundhog}, but they are designed for descriptive tasks~\cite{chen2023shikra, zellers2019recognition}, not for segmentation.
Another limitation of most existing models (except GRES models~\cite{liu2023gres}) is that they generate only one target per query, even if multiple targets are available in the image.

\begin{table}[ht]
    \centering
    \resizebox{\linewidth}{!}{%
    \begin{tabular}{l|c c|c c}
    \toprule
    \multirow{2}{*}{Paradigm} & \multicolumn{2}{c|}{Prompt} & \multicolumn{2}{c}{Target} \\
    & Text & Mask & Mask & Group \\
    \midrule
    Interactive segmentation~\cite{zou2024segment, kirillov2023segment} & $\circ$ & $\circ$ & \cmark & \xmark \\
    RES~\cite{hu2016segmentation, yan2023universal}, GRES~\cite{liu2023gres} & \cmark & \xmark & \cmark & $\circ$ \\
    LMM~\cite{liu2023visual, zhu2024minigpt, dai2023instructblip} & \cmark & \xmark & \xmark & \xmark \\
    Grounding LMM~\cite{lai2024lisa, rasheed2024glamm, zhang2024groundhog} & \cmark & $\circ$ & \cmark & $\circ$ \\
    \midrule
    ORES (Ours) & \cmark & \cmark & \cmark & \cmark \\
    \bottomrule
    \end{tabular}}
    \caption{\textbf{Comparison with existing paradigms.} Our omnimodal referring expression segmentation (ORES) task poses new challenges for all prior methods, including allowing mask-based visual prompts for reference visual entities and predicting a group of masks. \cmark: supported, $\circ$: partially supported, \xmark: unsupported.
    }
    \label{tab:comparison}
\end{table}

To address the ORES challenge, we adopt a simple yet effective approach that enables vision-language comprehension at the mask level. We \emph{extend segmentation foundation models~\cite{kirillov2023segment} with multimodal semantic understanding of segmentation masks} to leverage the strengths of both segmentation foundation models and LMMs:
Segmentation foundation models, such as SAM~\cite{kirillov2023segment}, benefit from large-scale training data with fine-grained mask annotations, but have limited semantic understanding of the produced masks; LMMs excel at language-based comprehension of visual inputs, but datasets that can train LMMs for pixel-level grounding are much smaller in scale~\cite{lai2024lisa}.
With such insight, we propose a framework, \textbf{Refer to Any Segmentation Mask Group (\ourmodel)}, to bridge the strengths of both sides. We first leverage the visual entity masking ability of segmentation models to propose a pool of candidate masks, which effectively covers the true targets, as we will show in Section~\ref{sec:expr-ablation}. Then, we introduce a \emph{mask-centric} LMM with enhanced semantic understanding of each visual entity encapsulated by candidate masks.

Specifically, our \ourmodel framework
a) employs a segmentation foundation model to propose candidate masks,
b) extracts semantic-rich visual feature maps with an ensemble of visual backbones~\cite{tong2024cambrian},
c) produces entity-level visual features by aggregating the features within each masked region to form \emph{mask tokens}, and
d) aligns mask tokens with a language model through visual instruction tuning~\cite{liu2023visual}. Notably, reference masks in vision-enriched prompts can be naturally converted into mask tokens as part of the input.
In this mask-centric formulation, each mask token is designed to encode one visual entity instead of a fixed-size image patch. This approach is more suitable for modeling the semantics of individual entities and their interactions.

ORES requires the model to output a group of target masks, which is essentially a set prediction problem~\cite{rezatofighi2017deepsetnet} and is known to pose difficulties in model optimization~\cite{sun2021rethinking}. To facilitate optimization, we adopt a \emph{non-autoregressive decoding}~\cite{carion2020end, vinyals2015pointer} procedure in \ourmodel. Instead of letting the model output the selected masks one by one autoregressively~\cite{lai2024lisa, xia2024gsva, zhang2024groundhog} in the prediction stage, we feed all candidate mask tokens into the model and learn to perform binary classification on each contextualized mask token to decide whether this candidate should be included in the group or not. With this design, we avoid directly predicting a sequence of mask embeddings~\cite{lai2024lisa, zhang2024groundhog} and effectively convert the set prediction problem into an easy-to-optimize per-mask binary classification problem.

To learn \ourmodel, we construct a large-scale instruction-tuning dataset \ourlargedata by automatically repurposing object-level annotations from existing datasets~\cite{lin2014microsoft, gupta2019lvis, krishna2017visual, yu2016modeling, liu2023gres}. Based on labeled categories, attributes, and relationships of objects, we create 2 million mask groups for visual instruction tuning.
Furthermore, in order to align \ourmodel with user preferences and evaluate its performance in real-world applications, we collect a high-quality mask grouping dataset \ourgooddata by requesting expert human annotators to propose meaningful visual entity groups and select the corresponding masks.

In summary, our main contributions include:
\begin{itemize}[leftmargin=*, noitemsep, nolistsep]
\item We introduce the omnimodal referring expression segmentation (ORES) task, which extends the classic RES and GRES tasks with vision-language prompts for more flexible and practical use cases.
\item We propose the Refer to Any Segmentation Mask Group (\ourmodel) framework to strengthen the semantic understanding of segmentation masks with a mask-centric LMM and produce mask groups for vision-language prompts.
\item We build a large-scale dataset \ourlargedata for instruction tuning of \ourmodel and curate \ourgooddata for alignment with human preferences and evaluation.
\item Empirical results demonstrate state-of-the-art performance of our solution on the newly proposed ORES dataset, as well as classic RES and GRES benchmarks.
\end{itemize}

\section{Related Work}
\label{sec:related}

\noindent\textbf{Referring expression segmentation (RES)} aims to segment one object at a time based on descriptions in natural language~\cite{hu2016segmentation, kazemzadeh2014referitgame, mao2016generation, yu2016modeling}. Earlier approaches focused on combining visual and language features~\cite{liu2017recurrent, li2018referring, chen2019see, ye2019cross, feng2021encoder, jing2021locate} and incorporating transformer models~\cite{wang2022cris, kim2022restr, yang2022lavt, yan2023universal}. Recent advancements~\cite{liu2023gres} have expanded the classic RES task to include multi-target and no-target queries, referred to as generalized RES (GRES). Building on this progress, our work further enhances GRES by enabling more effective and user-friendly interactions through the flexible integration of visual and textual inputs. 
 
\noindent\textbf{Large multimodal models (LMMs)} extend large language models (LLMs)~\cite{devlin2019bert, radford2018improving, touvron2023llama} with vision-language capabilities via visual instruction tuning~\cite{liu2023visual, zhu2024minigpt, dai2023instructblip}. Early LMMs are mainly based on CLIP~\cite{radford2021learning} patch-level visual features and show weaknesses in object-level comprehension and reasoning~\cite{tong2024eyes, tong2024cambrian, li2023evaluating, sun2024aligning}. LMMs can be equipped with grounding capabilities for generating bounding boxes~\cite{peng2024grounding, chen2023shikra, wang2023visionllm, pi2023detgpt, you2024ferret, li2024covlm} or segmentation masks~\cite{lai2024lisa, rasheed2024glamm, zhang2024groundhog, ren2024pixellm} via training on converted datasets with semantic-pixel alignment. Unlike prior LMMs, our \ourmodel is not trained for text generation, because text responses are unnecessary in the task and applications we consider (Figure~\ref{fig:teaser}), and high-quality mask groups are prioritized.

\noindent\textbf{Grounding LMMs} provide grounded vision-language understanding, and achieve state-of-the-art performance in RES and GRES~\cite{lai2024lisa, chng2024mask, zhang2024groundhog, rasheed2024glamm, xu2024ullava, chen2024sam4mllm, zhang2024psalm}. Among them, Groundhog~\cite{zhang2024groundhog} is most related to our work, which also performs RES by selecting from mask proposals. Our \ourmodel differs from Groundhog in these critical aspects: a) We adopt a non-autoregressive decoding procedure, outperforming the traditional autoregressive decoding used by Groundhog (Section~\ref{sec:expr-ablation}). b) We accept mask-based visual prompts for the complex ORES task, while for Groundhog, mask prompts are only effective in region description tasks.

\section{Refer to Any Segmentation Mask Group}
\label{sec:model}

\begin{figure*}[ht]
    \centering
    \vspace{-2mm}
    \includegraphics[width=\linewidth]{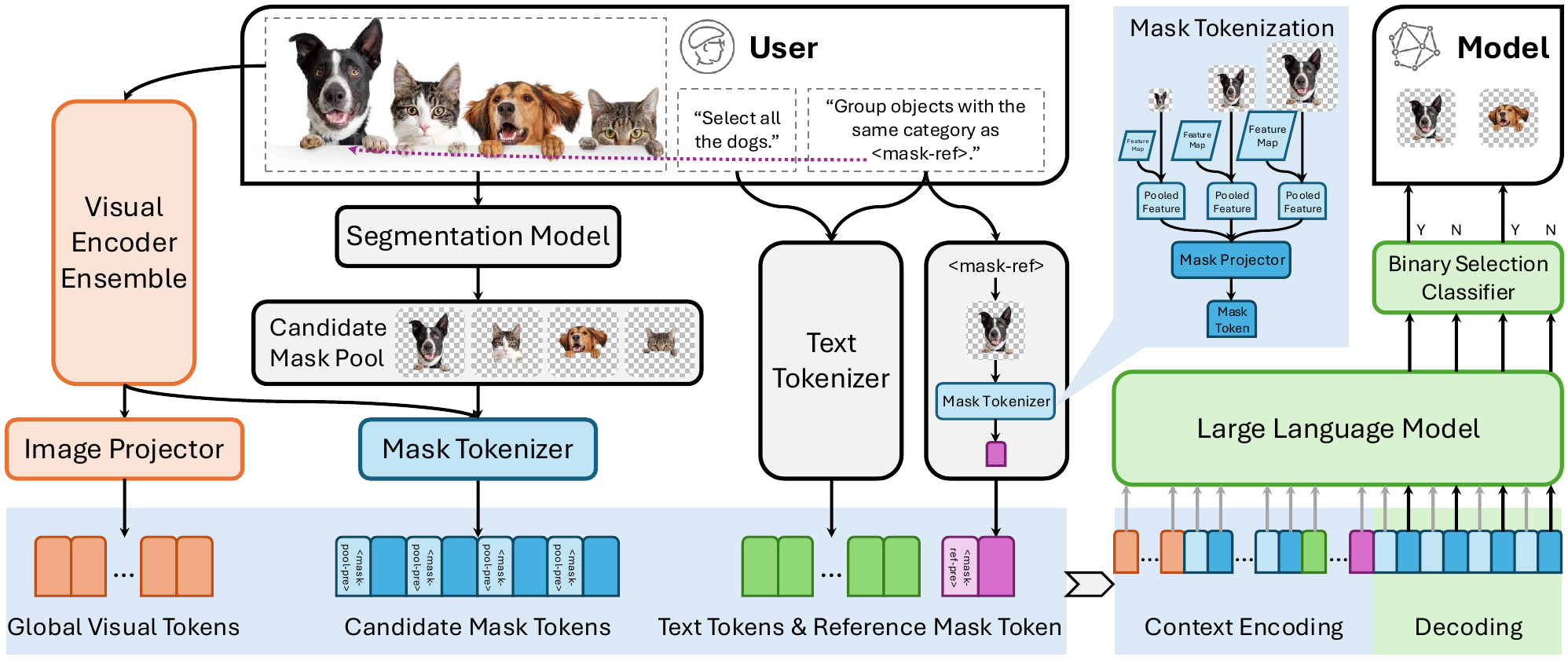}
    \caption{\textbf{Overview of our Refer to Any Segmentation Mask Group (\ourmodel) framework.} We extend LLaVA-1.5~\cite{liu2024improved} with a segmentation model, a visual encoder ensemble, mask tokenization, and a binary selection classifier for mask grouping. The decoding procedure of the LLM is non-autoregressive~\cite{carion2020end}, as the input tokens are given as candidate mask tokens rather than predicted from previous tokens.}
    \label{fig:model}
    \vspace{-2mm}
\end{figure*}

Extending the decoupling strategy in open-vocabulary segmentation~\cite{liang2023open, han2023open, yu2024towards}, our approach leverages a segmentation model to propose candidate masks for a given image. As the segmentation model does not directly comprehend the complex vision-language prompt, we design a \emph{mask-centric} large multimodal model (LMM) to address the new challenge of understanding and grouping these masks. Our proposed framework, \textbf{Refer to Any Segmentation Mask Group} (\textbf{\ourmodel}, illustrated in Figure~\ref{fig:model}), includes several specialized components: a segmentation model that proposes candidates, a mask projector that encodes mask features, a binary selection classifier that determines which masks to include, and a non-autoregressive decoding procedure for more effective model optimization.
We introduce the model designs in Section~\ref{sec:model-arch} and training procedure in Section~\ref{sec:model-train}.

\subsection{Architecture Designs}
\label{sec:model-arch}

\noindent\textbf{LMM as meta-architecture.} The widely adopted LMM architecture~\cite{liu2023visual} uses a CLIP visual encoder~\cite{radford2021learning} to extract features from a given image, and then maps the visual features into the language feature space via a lightweight \emph{image projector}. The converted visual tokens are concatenated with text tokens to form a sequence and fed into a large language model (LLM) to generate output responses autoregressively~\cite{radford2018improving}. Although LMMs have acquired strong image-level vision-language capabilities, they are not originally designed for tasks focusing on understanding fine-grained visual entities. Therefore, to perform the mask grouping task, we enhance LLaVA-1.5~\cite{liu2024improved} (finetuned from Vicuna-13B~\cite{vicuna2023}) with the ability to encode mask representations and select masks according to input prompts.

\noindent\textbf{Mask tokenization.} After segmenting the image into candidate masks, we tokenize the masks into individual elements for the LLM to understand. Given a segmentation mask (either proposed by segmentation models or specified by users) plus feature maps extracted by visual encoders from the entire image, we perform \emph{mask pooling} to aggregate visual features within the mask. More specifically, the mask is downsampled to the same spatial size as each visual feature map, and visual features within the downsampled mask are averaged to produce the mask-level feature. Then, a lightweight \emph{mask projector} converts the concatenated mask-level features into the language feature space, and finally we consider these converted features as mask tokens. This procedure is depicted in the mask tokenization block in Figure~\ref{fig:model}. Furthermore, we prepend a learnable special token \texttt{\textless mask-pool-pre\textgreater} to each token that corresponds to a candidate mask. This special token indicates that the following token will be a mask token converted from a continuous embedding of a mask in the pool of candidates. These mask tokens are concatenated with the global visual tokens and text tokens as the LLM inputs.

\noindent\textbf{Reference mask representation.} In the input, reference masks are encoded similarly to the candidate masks. For instance, in the prompt ``Select all objects with the same color as \texttt{\textless mask-ref\textgreater},'' the special token \texttt{\textless mask-ref\textgreater} would be replaced by the actual token of the reference mask specified by the user. We can reuse the mask tokenization for encoding candidate mask tokens, but one minor difference is that we prepend different special tokens to candidate mask tokens and reference mask tokens, in order to distinguish their roles. For candidate masks, we prepend \texttt{\textless mask-pool-pre\textgreater} to indicate that the next token will be an embedding for a candidate mask, and the model should interpret it as one possible choice for the mask grouping task. For reference masks, we prepend \texttt{\textless mask-ref-pre\textgreater} to indicate that the next token will be a reference mask embedding, which allows the model to extract related information for the mask grouping prompt.

\noindent\textbf{Visual feature ensemble.} Recent studies~\cite{tong2024eyes,tong2024cambrian} suggest that the CLIP visual encoder~\cite{radford2021learning} has several inherent weaknesses such as unlocalized features. Therefore, CLIP alone is not ideal for our mask grouping task, as we need \emph{localized features} to represent and distinguish different masks. Following Cambrian-1~\cite{tong2024cambrian}, we employ an ensemble of four visual encoders: CLIP~\cite{radford2021learning}, SigLIP~\cite{zhai2023sigmoid}, ConvNeXt-based CLIP~\cite{liu2022convnet, cherti2023reproducible}, and DINOv2~\cite{oquab2023dinov2}. In addition, we create a position-aware feature map from 2D sinusoidal position embeddings~\cite{dosovitskiy2021image} to explicitly encode positions. To tokenize each mask, after computing mask-pooled features from all encoders, we concatenate them all to produce an aggregated mask feature, and then use the mask projector to map the feature into a language-aligned token. Since the mask pooling operation is performed per feature map, we allow different resolutions for each visual encoder, which better preserves the original capabilities of each encoder.

\noindent\textbf{Mask group decoding.} Given a set of candidate masks, our model needs to predict a subset (group) of masks according to the prompt. For mask group prediction, a straightforward solution may be directly predicting continuous mask embeddings one by one in an autoregressive manner~\cite{lai2024lisa, xia2024gsva, zhang2024groundhog}. However, this is challenging because a) LLMs are originally trained to model a distribution over discrete tokens, and b) learning to predict an unordered set is inherently hard due to the instable bipartite matching between predictions and the ground truth~\cite{sun2021rethinking}. For consistency with the discrete nature of LLMs and effective model optimization, we reformulate the mask group prediction problem as a \emph{per-mask binary classification} problem (Figure~\ref{fig:model}). More specifically, we learn to make a binary prediction for each candidate mask to indicate whether it should be included in the mask group based on the input prompt. We first provide all the tokens that encode the context, and then feed the candidate mask tokens again to the LLM to capture their output hidden states. Leveraging the strong semantic understanding and reasoning capabilities of the LLM, the output hidden state can indicate whether a candidate mask is positively related to the user prompt. Finally, a learnable \emph{binary selection classifier} is applied on top of the hidden states to produce the binary predictions.

Note that our LLM decoding does not follow the autoregressive paradigm---the inference-time input to the LLM is fixed as the candidate mask tokens, instead of using the previous output tokens. As we will show in Section~\ref{sec:expr-ablation}, this simple and direct decoding strategy greatly outperforms traditional autoregressive decoding. Meanwhile, we can perform binary classification on all candidate masks in one pass to boost inference efficiency, while autoregressive generation can output only one token at a time.

\subsection{Multi-stage Training}
\label{sec:model-train}

\noindent\textbf{Training stages.} To efficiently train the model containing both pretrained weights and randomly initialized weights, we divide the training into two stages following the practice of LLaVA~\cite{liu2023visual}. The first stage is mask projector pretraining, during which we freeze all modules except the mask projector, as it is a new module that cannot inherit from LLaVA model weights.
We design a pretext task where we provide \emph{only mask tokens and text prompts} (without global visual tokens) to the LLM and let it predict the image-level description, which is similar to the original LLaVA pretraining task but uses mask tokens instead.
This task aligns mask tokens with the LLM, so that they can jointly produce image descriptions. We find it viable to \emph{caption images using only mask tokens} and the loss converges to a level similar to LLaVA pretraining, because mask tokens can capture and describe the major objects, which is enough to caption object-centric images. For this stage, we reuse the pretraining data from LLaVA, a set of image-caption pairs, and add SAM-generated~\cite{kirillov2023segment} masks to each image.

In the second stage, visual instruction tuning, we finetune all modules except the visual encoders for the mask grouping task. Given an image and a set of candidate masks, the model learns to predict the correct subset of masks based on the input vision-language prompt. The model can be further finetuned on task-specific data to adapt the model for greater specialization in downstream tasks like RES.

\noindent\textbf{Training objectives.} In the pretraining stage, the expected model outputs are text-only, so we can train the model using cross-entropy loss~\cite{liu2023visual, radford2018improving}. During the visual instruction tuning stage, the learning task transitions to mask grouping. Therefore, we change the training objective from maximizing the likelihood of the caption to optimizing for per-mask binary classification.
More specifically, we compute the mean binary cross-entropy loss, averaged over all candidate masks. Among numerous mask candidates, usually only a few should be selected in the group. Due to this imbalanced distribution of positive and negative samples, we assign a larger loss weight to positive candidates.

\section{Data for Model Training and Evaluation}
\label{sec:data}

To effectively train and evaluate \ourmodel, we build two datasets: \ourlargedata, which is a dataset containing 2 million samples automatically generated from existing datasets with object-level annotations, and \ourgooddata, which is a smaller, high-quality and diverse dataset annotated by human annotators.

\subsection{\ourlargedata: Data Repurposed for Visual Instruction Tuning}
Each training sample of the mask grouping task consists of an image, a set of candidate masks, a prompt (described by free-form text and optional reference masks), and a target mask group containing an arbitrary number of masks. No existing datasets provide all of these elements. To build \ourlargedata, we convert object-level annotations into the mask grouping format with templates. Table~\ref{tab:ourlargedata} summarizes the sources of each component in \ourlargedata. More details are in the supplementary material.

{
\setlength{\tabcolsep}{2pt}
\begin{table}[th]
    \centering
    \resizebox{\linewidth}{!}{%
    \begin{tabular}{l c c c c c c c c}
        \toprule
        & \multicolumn{4}{c}{w/o \texttt{\textless mask-ref\textgreater}} & \multicolumn{4}{c}{w/ \texttt{\textless mask-ref\textgreater}} \\
        \cmidrule(lr){2-5}\cmidrule(lr){6-9}
        Source & Cat. & Att. & Pos. & Free. & Cat. & Att. & Pos. & Free. \\
        \midrule
        MS-COCO+LVIS~\cite{lin2014microsoft, gupta2019lvis} & 166K & - & - & - & 166K & - & - & - \\
        VG~\cite{krishna2017visual} & 224K & 149K & 132K & - & 224K & 149K & 392K & 34K \\
        (G)RES~\cite{yu2016modeling, liu2023gres} & - & - & - & 474K & - & - & - & - \\
        \bottomrule
    \end{tabular}}
    \caption{\textbf{Composition of \ourlargedata.} We collect mask groups based on categories, attributes, positions, and other free-form descriptions by converting object-level annotations from MS-COCO~\cite{lin2014microsoft}, LVIS~\cite{gupta2019lvis}, Visual Genome~\cite{krishna2017visual}, and (generalized) referring expression segmentation datasets~\cite{yu2016modeling, liu2023gres}.}
    \label{tab:ourlargedata}
\end{table}
}

\begin{table*}[th]
    \centering
    \vspace{-2mm}
    \begin{tabular}{l c c c c c c}
        \toprule
        & \multicolumn{2}{c}{w/o \texttt{\textless mask-ref\textgreater}} & \multicolumn{2}{c}{w/ \texttt{\textless mask-ref\textgreater}} & \multicolumn{2}{c}{Overall} \\
        \cmidrule(lr){2-3}\cmidrule(lr){4-5}\cmidrule(lr){6-7}
        Model & gIoU & cIoU & gIoU & cIoU & gIoU & cIoU \\
        \midrule
        ReLA~\cite{liu2023gres} & 34.93 & 43.22 & - & - & - & - \\
        PSALM$_\text{1.3B}$~\cite{zhang2024psalm} & 36.92 & 37.33 & - & - & - & - \\
        GSVA$_\text{13B}$~\cite{xia2024gsva} & 41.98 & 49.55 & - & - & - & - \\
        \midrule
        \ourmodel$_\text{13B, SAM}$ (Ours) & 55.82 & 60.12 & 35.91 & 37.77 & 50.98 & 53.93 \\
        \ourmodel$_\text{13B, SAM, ORES-FT}$ (Ours) & \bf 66.71 & \bf 74.59 & \bf 58.72 & \bf 68.77 & \bf 64.77 & \bf 73.13 \\
        \bottomrule
    \end{tabular}
    \caption{\textbf{Results on our ORES dataset \ourgooddata.} Existing GRES models are unable to process reference masks as part of the input prompt (`-' in the table). Given text-only prompts, \ourmodel shows significantly stronger performance, which can be further improved by ORES finetuning. For LLM-based models, we mark the LLM scales in the subscript.}
    \vspace{-2mm}
    \label{tab:ourgooddata}
\end{table*}

\noindent\textbf{Category-based groups.} Given categorical annotations, we find same-category objects in each image, and form prompts in templates like ``Select all \texttt{\textless category\textgreater}'' or ``Segment everything of the same class as \texttt{\textless mask-ref\textgreater}.'' These groups originate from MS-COCO~\cite{lin2014microsoft}, LVIS~\cite{gupta2019lvis}, and Visual Genome~\cite{krishna2017visual}. LVIS uses the same images as MS-COCO but annotates more object categories with improved mask quality. Therefore, we merge MS-COCO and LVIS annotations before proposing category-based groups.

\noindent\textbf{Attribute-based groups.} Visual Genome~\cite{krishna2017visual} includes annotations for object attributes (\eg, colors, materials). Similar to category-based groups, we collect objects with the same attribute and formulate groups like ``Select all \texttt{\textless attribute\textgreater} objects'' or ``Find all the objects with the same attribute as \texttt{\textless mask-ref\textgreater} in the image.''

\noindent\textbf{Position-based groups.} The bounding box annotations provide positional information of objects. We form groups based on absolute positions (\eg, ``Locate all the items on the left side of the image.'') or relative positions (\eg, ``Find all the objects above \texttt{\textless mask-ref\textgreater}.'') by comparing the coordinates of the bounding boxes.

\begin{figure}[th]
    \centering
    \includegraphics[width=\linewidth]{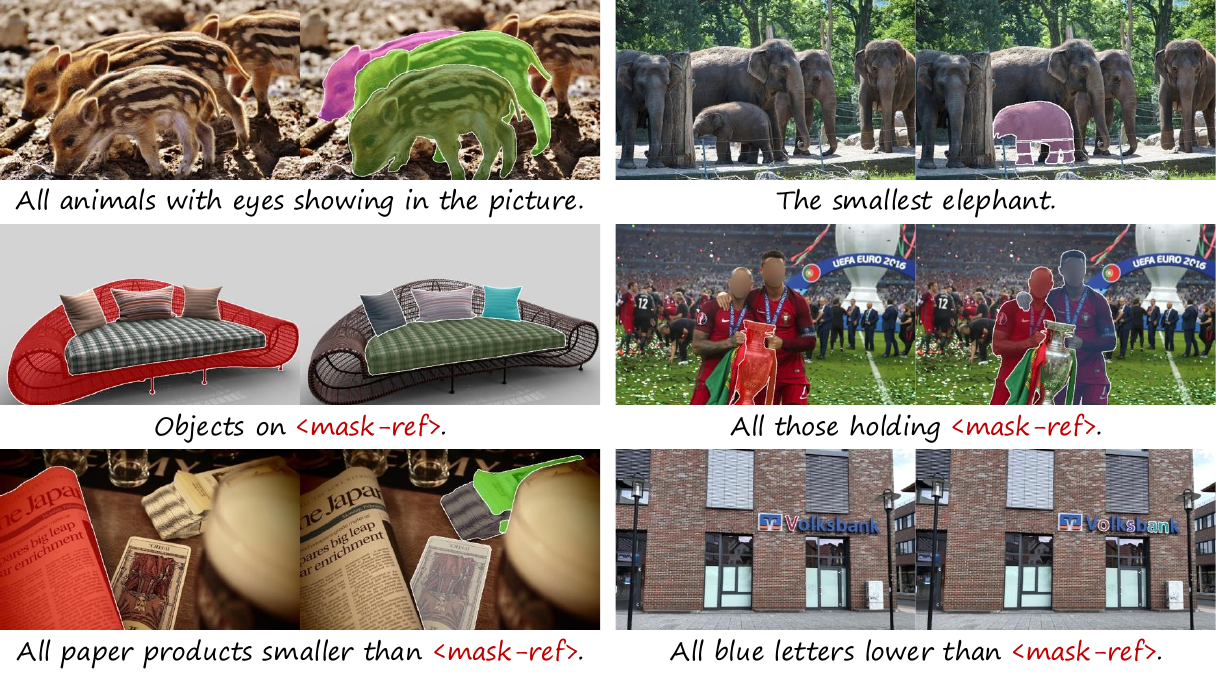}
    \caption{\textbf{Examples of \ourgooddata.} Diverse vision-language prompts are included, involving object categories, attributes, positions, comparisons, interactions, \etc. Best viewed on an electronic device with zoom-in functionality.}
    \vspace{-4mm}
    \label{fig:sample}
\end{figure}

\noindent\textbf{Other free-form prompts.} In addition to the mask grouping criteria introduced above, we include groups with diverse free-form descriptions. RES (RefCOCO, RefCOCO+, and RefCOCOg~\cite{yu2016modeling}) and GRES (gRefCOCO~\cite{liu2023gres}) datasets contain free-form phrases for localizing specific objects, which can be converted into text-only prompts and mask groups (\eg, ``Select the \texttt{\textless expression\textgreater} in the image.''). Visual Genome~\cite{krishna2017visual} contain annotations for object relationships. When there are multiple objects sharing the same relationship with the same subject, we group these objects (\eg, ``Select all objects that \texttt{\textless mask-ref\textgreater} \texttt{\textless relation\textgreater}.'').

\noindent\textbf{Avoiding data contamination.} The validation and test sets of RES/GRES datasets use images from MS-COCO and Visual Genome training data. To avoid data contamination, we exclude such images from \ourlargedata.

\subsection{\ourgooddata: Human-Annotated Data for Finetuning and Evaluation}
Although \ourlargedata is large enough for instruction tuning, its mask groups in pre-defined templates cannot cover all possible criteria that human users may use for grouping, and may introduce noises due to inaccurate annotations in the source data. To further improve and evaluate the generalizability of \ourmodel, we manually annotate a high-quality dataset \ourgooddata (visualized in Figure~\ref{fig:sample}). We start from EntitySeg~\cite{qi2023high}, an image segmentation dataset containing high-resolution images and category-agnostic masks. Notably, the images are from various sources, not only MS-COCO. Then, human annotators inspect the images and masks, and annotate several mask groups by proposing a reasonable vision-language prompt and labeling the IDs of masks that should be included. Our quality check ensures that the proposed mask groups are agreed upon by different users. In total, 100,299 mask groups are annotated. Finally, we split \ourgooddata into 96,697 samples over 18,368 images for finetuning, and 3,599 samples over 661 images for evaluation. 28\% of the samples include reference mask(s) in their prompts.

\section{Experiments}
\label{sec:expr}

\begin{table*}[th]
    \centering
    \vspace{-2mm}
    \begin{tabular}{l c c c c c c c c c}
        \toprule
        & \multicolumn{3}{c}{RefCOCO} & \multicolumn{3}{c}{RefCOCO+} & \multicolumn{2}{c}{RefCOCOg} & \multirow{2}{*}{Avg.} \\
        \cmidrule(lr){2-4}\cmidrule(lr){5-7}\cmidrule(lr){8-9}
        Model & val & testA & testB & val & testA & testB & val & test & \\
        \midrule
        ReLA~\cite{liu2023gres} & 73.8 & 76.5 & 70.2 & 66.0 & 71.0 & 57.7 & 65.0 & 66.0 & 68.3 \\
        LISA$_\text{13B, FT}$~\cite{lai2024lisa} & 74.9 & 79.1 & 72.3 & 65.1 & 70.8 & 58.1 & 67.9 & 70.6 & 69.9 \\
        MagNet~\cite{chng2024mask} & 76.6 & 78.3 & 72.2 & 68.1 & 73.6 & 61.8 & 67.8 & 69.3 & 71.0 \\
        Groundhog$_\text{7B}$~\cite{zhang2024groundhog} & 78.5 & 79.9 & 75.7 & 70.5 & 75.0 & 64.9 & 74.1 & 74.6 & 74.2 \\
        GSVA$_\text{13B, FT}$~\cite{xia2024gsva} & 79.2 & 81.7 & 77.1 & 70.3 & 73.8 & 63.6 & 75.7 & 77.0 & 74.8 \\
        GLaMM$_\text{7B, FT}$~\cite{rasheed2024glamm} & 79.5 & 83.2 & 76.9 & 72.6 & 78.7 & 64.6 & 74.2 & 74.9 & 75.6 \\
        u-LLaVA$_\text{7B}$~\cite{xu2024ullava} & 80.4 & 82.7 & 77.8 & 72.2 & 76.6 & 66.8 & 74.8 & 75.6 & 75.9 \\
        SAM4MLLM$_\text{8B}$~\cite{chen2024sam4mllm} & 79.8 & 82.7 & 74.7 & 74.6 & \bf 80.0 & 67.2 & 75.5 & 76.4 & 76.4 \\
        UNINEXT-H~\cite{yan2023universal} & 82.2 & 83.4 & 81.3 & 72.5 & 76.4 & 66.2 & 74.6 & 76.4 & 76.6 \\
        PSALM$_\text{1.3B}$~\cite{zhang2024psalm} & \bf 83.6 & \bf 84.7 & \bf 81.6 & 72.9 & 75.5 & 70.1 & 73.8 & 74.4 & 77.1 \\
        \midrule
        \ourmodel$_\text{13B, Co-DETR}$ (Ours) & 79.4 & 82.6 & 75.9 & 72.2 & 77.3 & 64.7 & 73.2 & 74.5 & 75.0 \\
        \ourmodel$_\text{13B, Co-DETR, RES-FT}$ (Ours) & 81.0 & 83.5 & 79.0 & \bf 75.1 & \bf 80.0 & \bf 70.3 & \bf 76.0 & \bf 77.5 & \bf 77.8 \\
        \bottomrule
    \end{tabular}
    \caption{\textbf{Results on referring expression segmentation (RES).} With Co-DETR, an instance segmentation model specialized for MS-COCO (retrained to avoid data leakage), we establish new state of the art in RES.
    Models that are finetuned again for RES after training on mixed data are labeled with the subscript $_\text{FT}$.}
    \label{tab:res}
    \vspace{-2mm}
\end{table*}

\begin{table*}[ht]
    \centering
    \resizebox{\linewidth}{!}{%
    \begin{tabular}{l c c c c c c c c c c}
        \toprule
        & \multicolumn{3}{c}{val} & \multicolumn{3}{c}{testA} & \multicolumn{3}{c}{testB} & Avg. \\
        \cmidrule(lr){2-4}\cmidrule(lr){5-7}\cmidrule(lr){8-10}
        Model & gIoU & cIoU & N-acc. & gIoU & cIoU & N-acc. & gIoU & cIoU & N-acc. & cIoU \\
        \midrule
        LAVT~\cite{yang2022lavt} & 58.40 & 57.64 & 49.32 & 65.90 & 65.32 & 49.25 & 55.83 & 55.04 & 48.46 & 59.33 \\
        ReLA~\cite{liu2023gres} & 63.60 & 62.42 & 56.37 & 70.03 & 69.26 & 59.02 & 61.02 & 59.88 & 58.40 & 63.85 \\
        LISA$_\text{13B, FT}$~\cite{lai2024lisa} & 65.24 & 63.96 & 57.49 & 69.99 & 71.00 & 55.43 & 62.11 & 62.29 & 56.34 & 65.75 \\
        HDC~\cite{luo2024hdc} & 68.28 & 65.42 & 63.38 & 72.52 & 71.60 & 65.29 & 63.85 & 62.79 & 60.68 & 66.60 \\
        GSVA$_\text{13B, FT}$~\cite{xia2024gsva} & 70.04 & 66.38 & 66.02 & 73.29 & 72.79 & \bf 64.72 & 65.45 & 63.20 & 62.47 & 67.46 \\
        SAM4MLLM$_\text{7B}$~\cite{chen2024sam4mllm} & 71.86 & 67.83 & 66.08 & 74.15 & 72.22 & 63.92 & 65.29 & 63.42 & 59.99 & 67.82 \\
        \midrule
        \ourmodel$_\text{13B, Co-DETR}$ (Ours) & 68.86 & 64.44 & 57.19 & 74.83 & 74.40 & 54.86 & 66.74 & 64.51 & 54.13 & 67.78 \\
        \ourmodel$_\text{13B, Co-DETR, GRES-FT}$ (Ours) & \bf 74.64 & \bf 70.48 & \bf 69.05 & \bf 77.45 & \bf 76.99 & 64.62 & \bf 69.42 & \bf 67.90 & \bf 62.92 & \bf 71.79 \\
        \bottomrule
    \end{tabular}}
    \caption{\textbf{Results on gRefCOCO for generalized referring expression segmentation (GRES).} Our approach achieves the best overall performance compared with baselines.}
    \label{tab:gres}
    \vspace{-2mm}
\end{table*}

In this section, we thoroughly test \ourmodel in various tasks, including omnimodal referring expression segmentation (ORES, Section~\ref{sec:expr-ours}), and classic and generalized referring expression segmentation (RES and GRES, Section~\ref{sec:expr-res}). We then analyze the candidate mask quality and design choices of \ourmodel (Section~\ref{sec:expr-ablation}), and finally demonstrate its applications (Section~\ref{sec:expr-demo}).
Following prior practice in RES, we mainly consider cumulative/generalized intersection over union (cIoU/gIoU) metrics; for GRES we also report the accuracy of identifying ``no-target'' samples (N-acc.)~\cite{liu2023gres}.

Due to limited space, we include a) additional results on comparison with SEEM~\cite{zou2024segment}, finetuning GSVA~\cite{xia2024gsva} on our data, and converting ORES visual prompts into language, b) ablation study on mask tokenization and LLM scales, and c) qualitative results in the supplementary material.

\subsection{Omnimodal RES}
\label{sec:expr-ours}
Our new ORES task poses new challenges to existing GRES models, because it uses mask-based visual prompts to describe relationships with reference entities. In Table~\ref{tab:ourgooddata}, we compare \ourmodel (both before and after finetuning on \ourgooddata) based on off-the-shelf SAM~\cite{kirillov2023segment} for proposing candidate masks, with state-of-the-art GRES models~\cite{liu2023gres, zhang2024psalm, xia2024gsva}. Unlike all GRES baselines, \ourmodel is able to accept reference masks and process all prompts in \ourgooddata, demonstrating a stronger prompting flexibility. Meanwhile, for text-only prompts, our results show significantly improved generalizability. Further finetuning \ourmodel on \ourgooddata brings more performance gains, especially enhancing the ability for understanding reference masks. Quantitative results are shown in Figure~\ref{fig:qual} in the supplementary material.

\subsection{RES and GRES}
\label{sec:expr-res}
Since our new task extends the classic and generalized referring expression segmentation (RES and GRES), \ourmodel can readily tackle both earlier tasks. We evaluate \ourmodel on RefCOCO, RefCOCO+, RefCOCOg, and gRefCOCO datasets~\cite{yu2016modeling, liu2023gres}, and compare it with prior state-of-the-art models~\cite{lai2024lisa, chng2024mask, zhang2024groundhog, rasheed2024glamm, xu2024ullava, chen2024sam4mllm, zhang2024psalm, yang2022lavt, luo2024hdc, liu2023gres, yan2023universal}. The segmentation modules in prior models are finetuned for the RES and GRES tasks. Therefore, to ensure a fair comparison, we employ a model trained for MS-COCO instance segmentation, Co-DETR~\cite{zong2023detrs}. To avoid data leakage, we retrain the Co-DETR instance segmentation model on MS-COCO excluding all RES/GRES validation and test images. As \ourmodel is not limited to a specific segmentation model, we can produce candidate masks with Co-DETR and seamlessly apply \ourmodel to its proposed masks.

As shown in Tables~\ref{tab:res} and \ref{tab:gres}, \ourmodel demonstrates competitive results on both RES and GRES. Following prior models that are also trained on mixed data~\cite{lai2024lisa, rasheed2024glamm, xia2024gsva}, we further finetune \ourmodel on RES/GRES training data to adapt \ourmodel for these tasks, which leads to state-of-the-art performance.

\subsection{Analysis and Ablation Study}
\label{sec:expr-ablation}
\noindent\textbf{Quality of candidate masks.} \ourmodel extends existing segmentation models with complex vision-language interactions. To validate that the segmentation models, Co-DETR and SAM, can propose candidate masks that sufficiently cover the referred targets, we analyze the proposed masks by selecting the ones with the hightest IoU with ground-truth masks and computing their cIoU on all three tasks. As shown in Table~\ref{tab:oracle}, \emph{even without finetuning on RES tasks}, the best candidates by Co-DETR and SAM achieve significantly better cIoU than existing RES models, and the proposals indeed include most of the true targets ($>$ 85 cIoU). By addressing the challenge of understanding and selecting high-quality candidates, \ourmodel achieves the best final results.

{
\setlength{\tabcolsep}{2pt}
\begin{table}[ht]
    \centering
    \vspace{-2mm}
    \resizebox{\linewidth}{!}{%
    \begin{tabular}{l c l c l c}
    \toprule
    \multicolumn{2}{c}{ORES} & \multicolumn{2}{c}{RES} & \multicolumn{2}{c}{GRES} \\
    \cmidrule(lr){1-2}\cmidrule(lr){3-4}\cmidrule(lr){5-6}
    Model & cIoU & Model & cIoU & Model & cIoU \\
    \midrule
    \multicolumn{6}{l}{\cellcolor{cellgray}\textit{Previous state of the art}} \\
    GSVA$_\text{13B}$ & 49.55 & PSALM$_\text{1.3B}$ & 77.1 & SAM4MLLM$_\text{7B}$ & 67.82 \\
    \multicolumn{6}{l}{\cellcolor{cellgray}\textit{Best candidates proposed by segmentation models in \ourmodel}} \\
    SAM$_\text{Oracle}$ & 86.39 & Co-DETR$_\text{Oracle}$ & 87.2 & Co-DETR$_\text{Oracle}$ & 87.60 \\
    \multicolumn{6}{l}{\cellcolor{cellgray}\textit{Final performance of \ourmodel (Ours)}} \\
    \ourmodel$_\text{13B, SAM}$ & 74.59 & \ourmodel$_\text{13B, Co-DETR}$ & 77.8 & \ourmodel$_\text{13B, Co-DETR}$ & 71.79 \\
    \bottomrule
    \end{tabular}}
    \caption{\textbf{Analysis of candidate mask quality.} ``Oracle'' denotes a setting where the ground-truth targets are known and the closest candidates are chosen. Compared with previous models, the best candidates proposed by the segmentation models in \ourmodel already obtain much higher mask quality, even without finetuning on RES tasks. Building upon the high-quality candidates, \ourmodel delivers the strongest final performance. }
    \label{tab:oracle}
    \vspace{-2mm}
\end{table}
}

Due to limited computation, models in the following ablation are only trained on 0.5M samples of \ourlargedata and tested on our ORES dataset \ourgooddata.

\noindent\textbf{Non-autoregressive \vs autoregressive decoding.} \ourmodel uses a simple yet effective decoding strategy, where the LLM inputs are directly from the candidate mask tokens rather than the previously predicted tokens. We compare our non-autoregressive formulation with the traditional autoregressive paradigm adopted by previous LMMs~\cite{lai2024lisa, xia2024gsva, zhang2024groundhog}. The autoregressive baseline learns to predict continuous mask embeddings of the selected masks in a sequential manner, and we collect the candidate masks whose embeddings are closest to these predicted embeddings as the output. Table~\ref{tab:ar} shows that our non-autoregressive formulation improves performance and enables more efficient inference.

\begin{table}[ht]
    \centering
    \begin{tabular}{l c c}
        \toprule
        Decoding Paradigm & cIoU$\uparrow$ & Latency$\downarrow$ \\
        \midrule
        Autoregressive & 45.34 & 2.13 \\
        \cellcolor{cellgray}Non-Autoregressive & \cellcolor{cellgray}\bf 53.75 & \cellcolor{cellgray}\bf 0.56 \\
        \bottomrule
    \end{tabular}
    \caption{\textbf{Comparison between decoding paradigms.} Our non-autoregressive formulation leads to more effective training and more efficient inference.}
    \label{tab:ar}
    \vspace{-2mm}
\end{table}

\noindent\textbf{Visual encoders.} We use four visual encoders as a feature ensemble for mask tokenization. In Table~\ref{tab:encoder}, we compare \ourmodel with its variants that encode mask tokens with a single encoder, as well as the previously best model GSVA~\cite{xia2024gsva}. \emph{Even with one single encoder}, \ourmodel outperforms GSVA. Combining all four encoders leads to the best results.

{
\setlength{\tabcolsep}{2pt}
\begin{table}[ht]
    \centering
    \vspace{-2mm}
    \resizebox{\linewidth}{!}{%
    \begin{tabular}{l c c c}
        \toprule
        Model & w/o \scriptsize\texttt{\textless mask-ref\textgreater} & w/ \scriptsize\texttt{\textless mask-ref\textgreater} & Overall cIoU \\
        \midrule
        GSVA$_\text{13B}$~\cite{xia2024gsva} & 49.55 & - & - \\
        \midrule
        \ourmodel$_\text{13B, CLIP, SAM}$ & \bf 58.13 & 37.61 & 52.44 \\
        \ourmodel$_\text{13B, ConvCLIP, SAM}$ & 56.83 & 44.06 & 53.53 \\
        \ourmodel$_\text{13B, SigLIP, SAM}$ & 54.24 & 32.09 & 48.07 \\
        \ourmodel$_\text{13B, DINOv2, SAM}$ & 57.40 & 21.70 & 47.71 \\
        \cellcolor{cellgray}\ourmodel$_\text{13B, Ensemble, SAM}$ & \cellcolor{cellgray}57.73 & \cellcolor{cellgray}\bf 44.47 & \cellcolor{cellgray} \bf 53.75 \\
        \bottomrule
    \end{tabular}}
    \caption{\textbf{Comparison of \ourmodel with different visual encoders.} Our ensemble of four visual encoders yields the best visual features for mask tokenization.}
    \label{tab:encoder}
    \vspace{-2mm}
\end{table}
}

\subsection{Applications}
\label{sec:expr-demo}
\begin{figure}[ht]
    \centering
    \includegraphics[width=\linewidth]{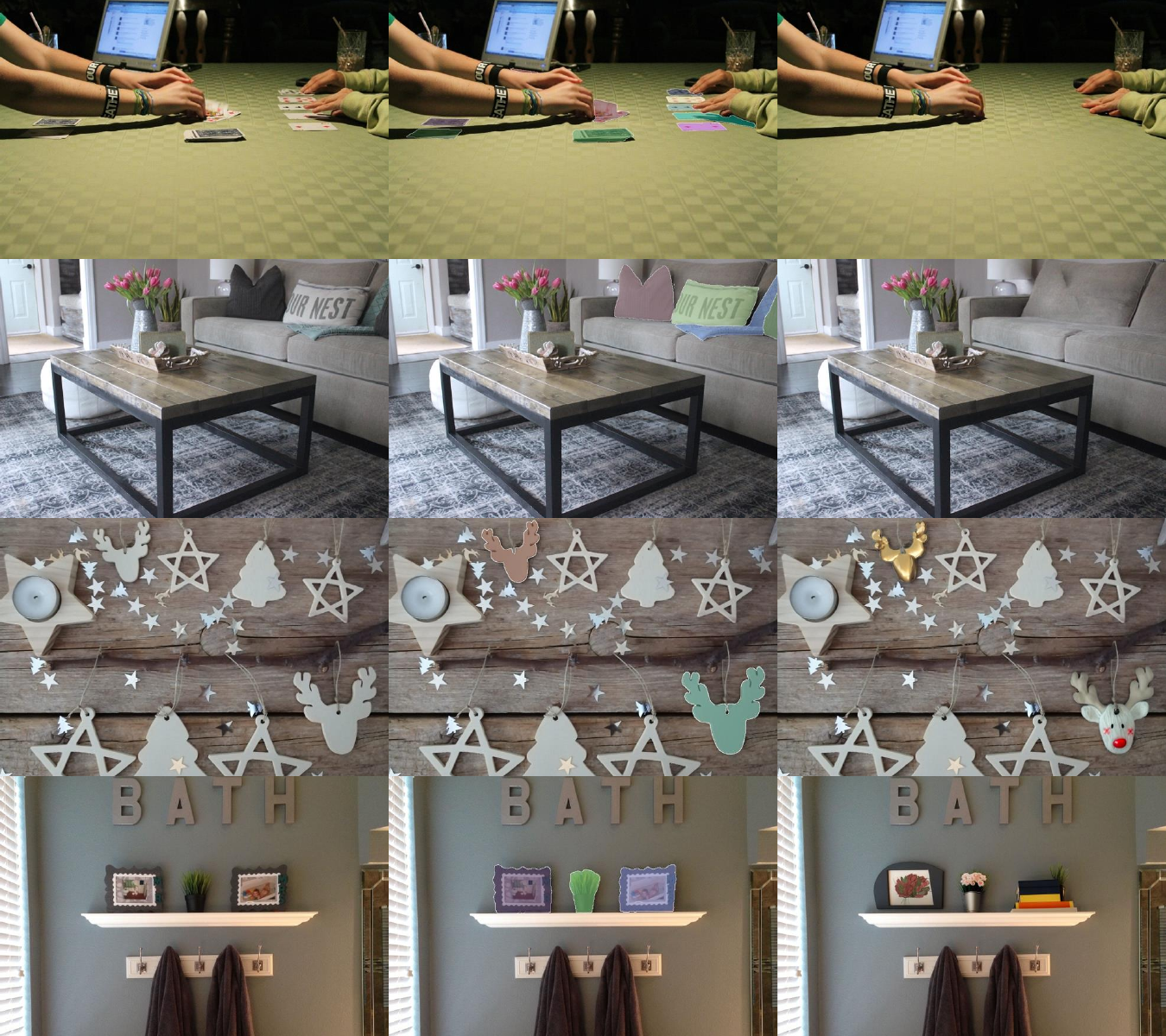}
    \caption{\textbf{Fine-grained image content manipulation enabled by our approach.} In each row we visualize the original image, the predicted segmentation masks, and the object removal (first two rows) or editing (last two rows) results. Best viewed on an electronic device with zoom-in functionality.}
    \label{fig:demo}
    \vspace{-4mm}
\end{figure}

By addressing the ORES task, \ourmodel improves a range of applications requiring fine-grained localization of multiple visual entities. As shown in Figure~\ref{fig:demo}, the predicted segmentation mask groups can be seamlessly integrated with generative models (\eg, Adobe Photoshop Generative Fill~\cite{adobephotoshop}) to \emph{remove or edit multiple targets} conveniently and efficiently.

\section{Conclusion}
\label{sec:con}
This work introduces a novel task, omnimodal referring expression segmentation (ORES), which extends RES with more sophisticated interactions through visual and textual prompts. We present a simple yet effective solution, \ourmodel, to achieve complex multimodal comprehension of segmentation masks. We demonstrate state-of-the-art performance compared to various baselines, not only in ORES but also in existing tasks (\ie, RES and GRES).

Future directions include leveraging the potentials of LMMs to enable capabilities such as generating textual justifications for predicted mask groups and supporting multi-round interactions. Furthermore, we plan to explore improving the synergy between the segmentation model and the LMM, along with developing compact models variants tailored specifically for ORES, based on computation-efficient LLMs.

\paragraph{Acknowledgments.} This work was supported in part by NSF Grant 2106825 and NIFA Award 2020-67021-32799. This work used computational resources, including the NCSA Delta and DeltaAI supercomputers through allocations CIS230012, CIS230013, CIS240133, and CIS240428 from the Advanced Cyberinfrastructure Coordination Ecosystem: Services \& Support (ACCESS) program, as well as the TACC Frontera supercomputer, Amazon Web Services, and OpenAI API through the National Artificial Intelligence Research Resource (NAIRR) Pilot.

{
    \small
    \bibliographystyle{ieeenat_fullname}
    \bibliography{main}
}

\clearpage
\appendix
\maketitlesupplementary
\setcounter{section}{0}
\setcounter{figure}{0}
\setcounter{table}{0}
\renewcommand{\thesection}{\Alph{section}}
\renewcommand{\thefigure}{\Alph{figure}}
\renewcommand{\thetable}{\Alph{table}}

\noindent In this appendix, we provide additional details on the implementation of our model \ourmodel (Section~\ref{sec:supp-impl}) and our datasets \ourlargedata (Section~\ref{sec:supp-ourlargedata}) and \ourgooddata (Section~\ref{sec:supp-ourgooddata}). Furthermore, we include additional experiments (Section~\ref{sec:supp-expr}), ablation study (Section~\ref{sec:supp-ablation}), and qualitative results (Section~\ref{sec:supp-qual}).

\section{Implementation Details}
\label{sec:supp-impl}
\noindent\textbf{Visual encoder ensemble.} Following Cambrian-1~\cite{tong2024cambrian}, we use four visual encoders: OpenAI CLIP ViT-L/14@336~\cite{radford2021learning}, OpenCLIP ConvNeXt-XXL@1024~\cite{liu2022convnet, cherti2023reproducible}, SigLIP ViT-SO400M/14@384~\cite{zhai2023sigmoid}, and DINOv2 ViT-L/14@518~\cite{oquab2023dinov2}. In addition, we provide 2D sinusoidal position embeddings~\cite{dosovitskiy2021image} of shape $32\times 32$ and treat them as visual features produced by a fifth visual encoder. All input images are padded to an aspect ratio of $1:1$, resized to the input size required by each encoder (up to $1,024 \times 1,024$), and fed into each encoder. All visual encoders are frozen during the entire training process.

\noindent\textbf{Mask projector and its pretraining.} We initialize \ourmodel with weights from LLaVA-1.5-13B~\cite{liu2024improved}. The mask projector is a two-layer multilayer perceptron (MLP) that projects the concatenated mask-level visual features to the language model space. As a new module, the mask projector is randomly initialized. Before training the whole \ourmodel model, we first pretrain the mask projector on the LLaVA-Pretrain dataset~\cite{liu2023visual, liu2024improved} with a modified pretext task. We use SAM~\cite{kirillov2023segment} to generate a set of masks per image and replace the original image tokens with our mask tokens for the image captioning objective. To correctly understand and describe a given image, the model needs to align the mask tokens with the LLM feature space. During the pretraining stage, we set the batch size to 128 and set the base learning rate to $1\times 10^{-3}$. We train on LLaVA-Pretrain for 1 epoch.

\noindent\textbf{Visual instruction tuning.} After pretraining the mask projector, the entire \ourmodel model (except the visual encoders) is trained in the visual instruction tuning stage. A binary selection classifier (two-layer MLP) is randomly initialized. Then, we minimize a binary cross-entropy loss. Due to the imbalanced distribution of positive/negative samples (usually only a few masks should be selected from a large pool of candidate masks), we assign a loss weight of $5.0$ to positive candidates. During the visual instruction tuning stage, we set the batch size to 128 and set the base learning rate to $2\times 10^{-5}$. We train on \ourlargedata for 1 epoch.

\noindent\textbf{Further finetuning.} For improved performance on specialized tasks (ORES, RES, and GRES), we further finetune \ourmodel on these tasks separately. We set the batch size to 64 and use the same base learning rate as instruction tuning. Due to different data scales, we finetune \ourmodel on ORES or GRES for 4 epochs, and finetune \ourmodel on RES for 2 epochs.

\noindent\textbf{Optimization and computation.} Following Vicuna~\cite{vicuna2023} and LLaVA~\cite{liu2023visual}, we use a cosine learning rate schedule with warm-up in each training stage. The optimizer is Adam~\cite{kingma2015adam} with zero weight decay. All of our training is performed on 8 NVIDIA A100-80GB GPUs. The pretraining stage requires about 4 hours. The visual instruction tuning stage on \ourlargedata requires about 1.5 days. Further finetuning for ORES, RES, or GRES requires another 1.5 days.

\section{Construction of \ourlargedata}
\label{sec:supp-ourlargedata}
\ourlargedata is converted from object-level annotations of existing image datasets. The sources of \ourlargedata are detailed as follows.

\noindent\textbf{MS-COCO~\cite{lin2014microsoft} and LVIS~\cite{gupta2019lvis}.} Since LVIS uses the same images as MS-COCO, we merge their annotations by combining instances with overlapping masks. For each image, we find object categories with at least 2 object annotations and create a category-based mask group with or without reference masks.

\noindent\textbf{Visual Genome~\cite{krishna2017visual}.} Because mask annotations are not provided by Visual Genome, we first use SAM~\cite{kirillov2023segment} to produce segmentation masks based on bounding box annotations and filter low-quality masks. We create category-based mask groups and attribute-based mask groups, similar to MS-COCO and LVIS. Furthermore, we compare the coordinates of bounding boxes to decide if an object is on the left side of, on the right side of, on the top of, or at the bottom of the entire image or another object, and then produce position-based mask groups with or without reference masks. In addition, Visual Genome provides annotations of relationships, which are triplets of (subject, relationship, object). In each image, we find triplets with a) the same subject and the same relationship but different objects, or b) the same object and the same relationship but different subjects, and formulate mask groups accordingly.

\noindent\textbf{RES~\cite{yu2016modeling} and GRES~\cite{liu2023gres}.} The RES datasets, including RefCOCO, RefCOCO+, and RefCOCOg, provide correspondences between a referring expression and an object, which can be directly converted into a single-mask group. The GRES dataset, gRefCOCO, contains referring expressions and their target object sets, and they can be converted into mask groups including a varying number (zero, one, or more than one) of masks. To avoid data contamination, we exclude images that are used for RES/GRES validation or test sets from the entire \ourlargedata dataset.

\begin{figure}[th]
    \centering
    \includegraphics[width=\linewidth]{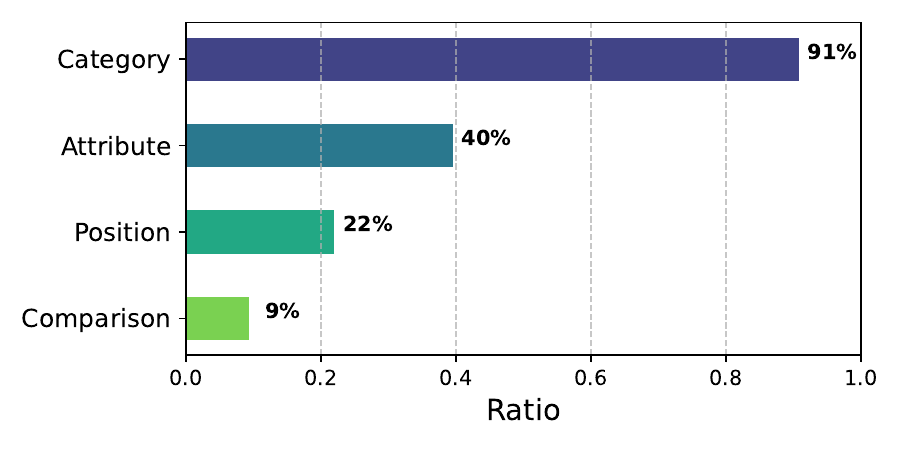}
    \caption{\textbf{Prompt type distribution in \ourgooddata.} A grouping criterion may involve the categories, the attributes, the absolute or relative positions, the cross-entity comparisons, and even their combination.}
    \label{fig:ourgooddata}
\end{figure}

\section{Statistics of \ourgooddata}
\label{sec:supp-ourgooddata}
\ourgooddata extends the existing mask annotations in EntitySeg~\cite{qi2023high} with vision-language prompts and mask groups. Human annotators are encouraged to propose creative and meaningful entity groups, so the prompts are very diverse and difficult to categorize. Nevertheless, we provide some statistics for better understanding of \ourgooddata: 28\% of the samples include reference masks in the prompts, and the other 72\% do not contain reference masks. In Figure~\ref{fig:ourgooddata}, we visualize the distribution of the prompts based on their grouping criterion. Note that each prompt may be labeled with multiple types. For example, the prompt ``All paper products smaller than \texttt{\textless mask-ref\textgreater}'' simultaneously involves a category (``paper product''), an attribute (``small''), and a comparison (``smaller than \texttt{\textless mask-ref\textgreater}'').

\section{Additional Experiment Results}
\label{sec:supp-expr}

\begin{figure}[th]
    \centering
    \includegraphics[width=\linewidth]{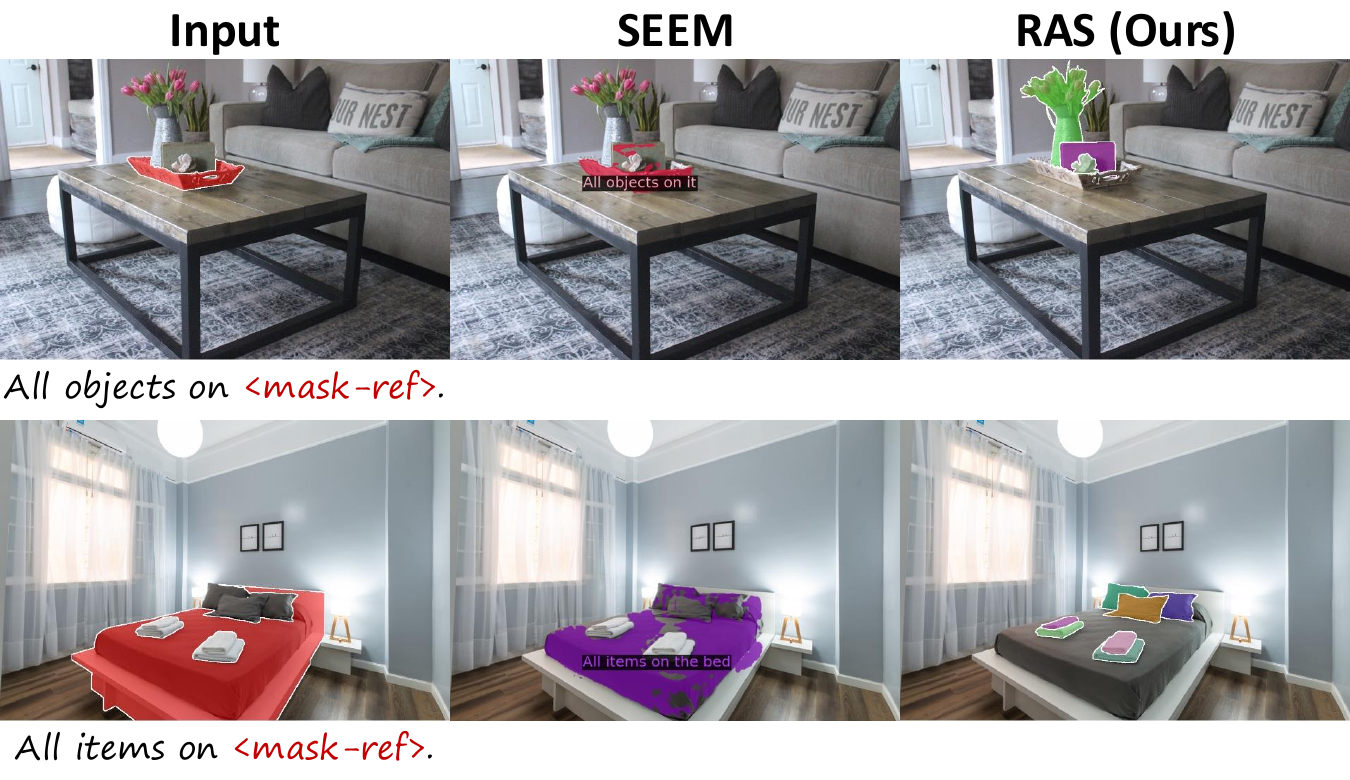}
    \caption{\textbf{SEEM, a representative interactive segmentation model, fails in our ORES task.} Instead of understanding the relationship (\eg, ``on the reference entity'') specified by the vision-language prompt, SEEM~\cite{zou2024segment} simply produces a mask that overlaps with the visual prompt. In contrast, our proposed \ourmodel model can correctly understand the vision-language prompt.}
    \label{fig:seem}
\end{figure}

\noindent\textbf{SEEM on ORES.} As introduced in the main paper, though some interactive segmentation models such as SEEM~\cite{zou2024segment} are able to take text and visual prompts simultaneously, their visual prompts can only be directly used for locating the target object. In contrast, visual prompts in ORES are often for a reference object that has a certain relationship with the target. In Figure~\ref{fig:seem}, we visualize examples of prompting SEEM with both text and visual prompts and compare the results with our model \ourmodel. SEEM outputs masks directly corresponding to the visual prompt, instead of correctly understanding the mixed prompt as required by the ORES task. In contrast, our model \ourmodel successfully selects the correct group of masks.

\noindent\textbf{Finetuning GSVA on our data.} To understand the impact of training data, we finetune GSVA~\cite{xia2024gsva}, the previously best GRES model, on our data and evaluate its ORES performance on \ourgooddata. As shown in Table~\ref{tab:gsva}, finetuning GSVA on samples from \ourlargedata does not yield better performance than its original data recipe, \ie, finetuning on GRES data, and is significantly worse than \ourmodel trained on the same data. Finetuning \ourmodel on the training split of \ourgooddata also leads to better ORES performance than GSVA. Note that training on MaskGroups-2M does not necessarily provide an advantage for performance on MaskGroups-HQ due to the domain gap: The samples in \ourlargedata are constructed from fixed templates, while the samples from \ourgooddata are written by human annotators in any free form. Therefore, the stronger performance of our model \ourmodel should be attributed more to its model design.

\begin{table}[t!]
    \centering
    \resizebox{\linewidth}{!}{%
    \begin{tabular}{l l c c}
        \toprule
        & & \multicolumn{2}{c}{w/o \texttt{\textless mask-ref\textgreater}} \\
        \cmidrule(lr){3-4}
        Model & Data & gIoU & cIoU \\
        \midrule
        GSVA$_\text{13B}$~\cite{xia2024gsva} & GRES (original) & 41.98 & 49.55 \\
        GSVA$_\text{13B}$~\cite{xia2024gsva} & 0.5M of \ourlargedata & 41.21 & 36.40 \\
        GSVA$_\text{13B}$~\cite{xia2024gsva} & \ourgooddata & 56.79 & 70.11 \\
        \midrule
        \ourmodel$_\text{13B, SAM}$ (Ours) & 0.5M of \ourlargedata & 54.76 & 57.73 \\
        \ourmodel$_\text{13B, SAM, ORES-FT}$ (Ours) & \ourgooddata & \bf 66.71 & \bf 74.59 \\
        \bottomrule
    \end{tabular}}
    \caption{\textbf{Results of finetuning GSVA on our data.} Finetuning GSVA~\cite{xia2024gsva}, the previously best GRES model, on samples of \ourlargedata, does not achieve better ORES performance than the GSVA model trained with its original data recipe. When finetuned on the training samples of \ourgooddata, \ourmodel significantly outperforms GSVA in the ORES task.}
    \label{tab:gsva}
\end{table}

\noindent\textbf{Converting visual prompts into language.} In the main paper, we have discussed the limitations of existing GRES models~\cite{liu2023gres, xia2024gsva, zhang2024psalm}: They cannot take visual prompts that represent reference entities as inputs, and therefore cannot process all samples in the ORES task (Table~\ref{tab:ourgooddata}). One may argue that visual prompts in ORES can be replaced by text prompts (\eg, ``Locate all pillows on \texttt{\textless mask-ref\textgreater}'' $\rightarrow$ ``Locate all pillows on the bed'', Figure~\ref{fig:teaser}). However, when the scene is complex and involves multiple semantically similar objects, visual prompts can hardly be clearly and concisely ``translated'' into language. To investigate this discrepancy between visual prompts and text prompts, we manually convert \texttt{\textless mask-ref\textgreater} into language for 200 samples in \ourgooddata, and test GRES models and our \ourmodel on these samples. As shown in Table~\ref{tab:translation}, visual prompts are better perceived by \ourmodel, indicating that such visual prompts are necessary to guide the model in accurately locating the target entities that are related to the reference entity. When provided with the same pure-text prompts, despite the increased complexity of the converted prompts, \ourmodel still outperforms the existing GRES models.

\begin{table}[t!]
    \centering
    \resizebox{\linewidth}{!}{%
    \begin{tabular}{l l c c}
        \toprule
        & & \multicolumn{2}{c}{w/ \texttt{\textless mask-ref\textgreater}} \\
        \cmidrule(lr){3-4}
        Prompt & Model & gIoU & cIoU \\
        \midrule
        \multirow{5}{*}{Text + Converted \texttt{\textless mask-ref\textgreater}} & ReLA~\cite{liu2023gres} & 21.15 & 24.14 \\
        & PSALM$_\text{1.3B}$~\cite{zhang2024psalm} & 24.68 & 24.19 \\
        & GSVA$_\text{13B}$~\cite{xia2024gsva} & 22.66 & 25.10 \\
        & \ourmodel$_\text{13B, SAM}$ (Ours) & 27.13 & 27.74 \\
        & \ourmodel$_\text{13B, SAM, ORES-FT}$ (Ours) & 43.76 & 47.80 \\
        \midrule
        \multirow{2}{*}{Text + Visual \texttt{\textless mask-ref\textgreater}} & \ourmodel$_\text{13B, SAM}$ (Ours) & 35.91 & 37.77  \\
        & \ourmodel$_\text{13B, SAM, ORES-FT}$ (Ours) & \bf 58.72 & \bf 68.77 \\
        \bottomrule
    \end{tabular}}
    \caption{\textbf{Results of converting visual prompts into language.} We manually translate visual prompts for reference entities into language (\eg, ``Locate all pillows on \texttt{\textless mask-ref\textgreater}'' $\rightarrow$ ``Locate all pillows on the bed,'' see Figure~\ref{fig:teaser}), and test multiple GRES models and our \ourmodel model on the converted prompts.
    The original visual prompts lead to better performance than the converted prompts, demonstrating that visual prompting is necessary in referring expression segmentation.
    When provided with pure-text prompts, our model \ourmodel still outperforms all prior GRES models. The subscript $_\text{ORES-FT}$ means evaluation of \ourmodel that is further finetuned on the original training set (not including the converted prompts) of \ourgooddata.}
    \label{tab:translation}
\end{table}

\section{Additional Ablation Study}
\label{sec:supp-ablation}

\noindent\textbf{Special tokens in mask tokenization.} In \ourmodel, we prepend a learnable special token \texttt{\textless mask-pool-pre\textgreater} to each candidate mask token and prepend a \texttt{\textless mask-ref-pre\textgreater} token to each reference mask token. These special tokens indicate the different roles of the following tokens. In Table~\ref{tab:token}, we compare \ourmodel with two variants: The first variant does not prepend any special tokens to the mask tokens, and the second variant prepends the same token to both candidate mask tokens and reference mask tokens. Using two different special tokens in mask tokenization achieves the best performance.

\begin{table}[t!]
    \centering
    \resizebox{\linewidth}{!}{%
    \begin{tabular}{l c c c}
        \toprule
        Special tokens & w/o \scriptsize\texttt{\textless mask-ref\textgreater} & w/ \scriptsize\texttt{\textless mask-ref\textgreater} & Overall cIoU \\
        \midrule
        No \texttt{\textless pre\textgreater} tokens & 55.61 & 34.98 & 50.13 \\
        Same \texttt{\textless pre\textgreater} tokens & 54.68 & 32.37 & 48.49 \\
        \cellcolor{cellgray}Different \texttt{\textless pre\textgreater} tokens & \cellcolor{cellgray}\bf 57.73 & \cellcolor{cellgray}\bf 44.47 & \cellcolor{cellgray} \bf 53.75 \\
        \bottomrule
    \end{tabular}}
    \caption{\textbf{Comparison of \ourmodel with different special tokens prepended to mask tokens.} Prepending \texttt{\textless mask-pool-pre\textgreater} to candidate mask tokens and \texttt{\textless mask-ref-pre\textgreater} to reference mask tokens leads to the best result. All models are trained on the same 0.5M samples from \ourlargedata and evaluated on \ourgooddata.}
    \label{tab:token}
\end{table}

\noindent\textbf{LMM scales.} In the main paper, we report the results of training our model \ourmodel based on LLaVA-1.5-13B~\cite{liu2024improved}, which originates from Vicuna-13B~\cite{vicuna2023}. In principle, \ourmodel can be built on other LLMs of different parameter scales. As an example, we train another \ourmodel based on LLaVA-1.5-7B. The model performance is summarized in Table~\ref{tab:llm}.

\begin{table}[t!]
    \centering
    \begin{tabular}{l c c c}
        \toprule
        Model & ORES & RES & GRES \\
        \midrule
        \ourmodel$_\text{7B, SAM / Co-DETR}$ & 52.19 & 73.7 & 67.30 \\
        \cellcolor{cellgray}\ourmodel$_\text{13B, SAM / Co-DETR}$ & \cellcolor{cellgray}\bf 53.93 & \cellcolor{cellgray}\bf 75.0 & \cellcolor{cellgray}\bf 67.78 \\
        \bottomrule
    \end{tabular}
    \caption{\textbf{Comparison of \ourmodel with different LLM scales.} The larger 13B LLM leads to a stronger performance on all tasks. The metric is the overall cIoU. We use SAM as the mask proposal model in ORES, and use Co-DETR in RES and GRES, consistent with the main results in Tables~\ref{tab:ourgooddata}, \ref{tab:res}, and \ref{tab:gres}.}
    \label{tab:llm}
\end{table}

\section{Additional Qualitative Results}
\label{sec:supp-qual}

\begin{figure}[ht]
    \centering
    \includegraphics[width=\linewidth]{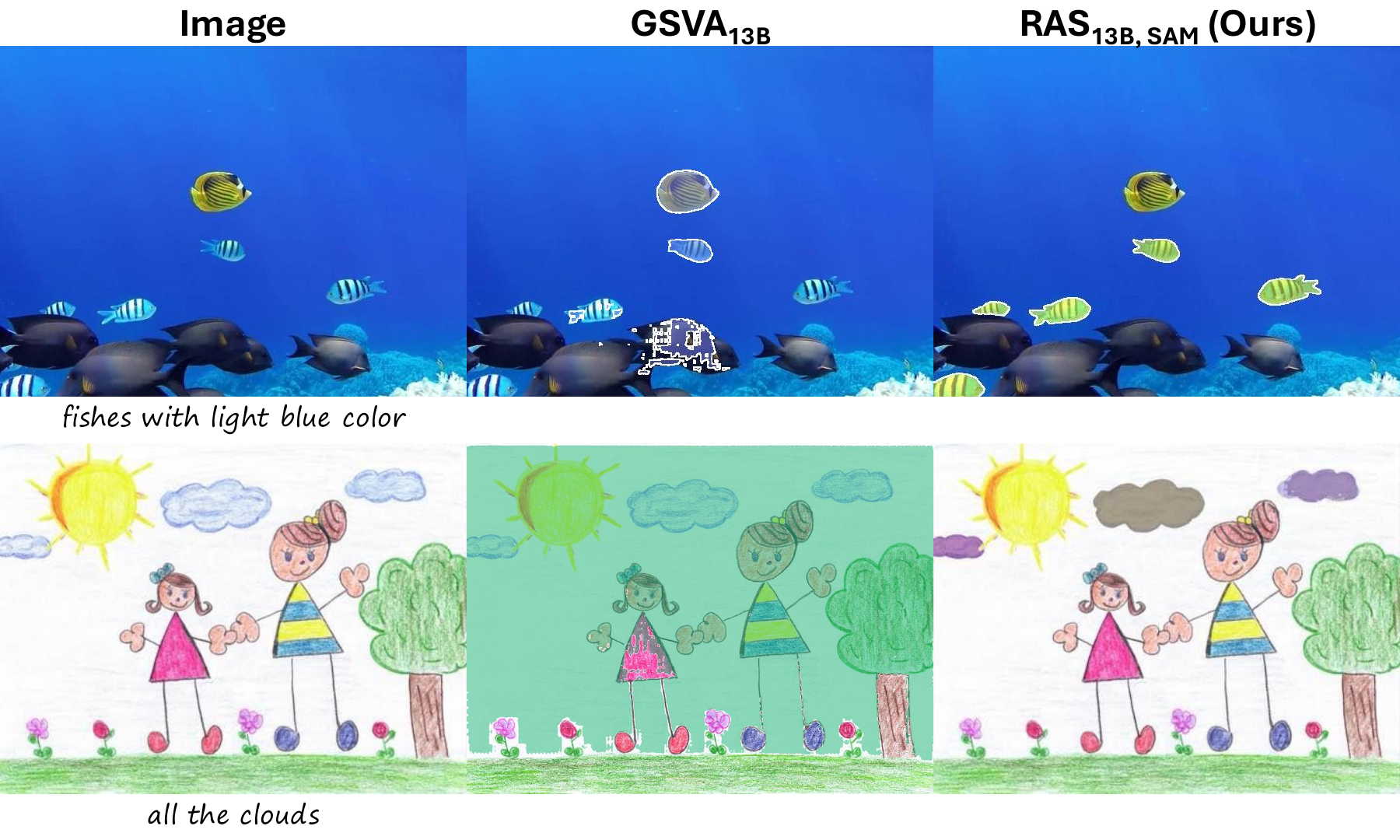}
    \caption{\textbf{Qualitative comparison on OOD examples.} Our RAS framework generalizes better to novel image domains, such as under-water images and cartoon-style images.}
    \label{fig:ood}
\end{figure}

Our RAS shows strong generalization beyond MS-COCO benchmarks, where prior works primarily focus. As shown in Figure~\ref{fig:ood}, our model outperforms GSVA on out-of-distribution (OOD) images. This is achieved by decoupling mask generation and selection, allowing RAS to leverage strong generalization capabilities of SAM.

\begin{figure*}[ht]
    \centering
    \includegraphics[width=\linewidth]{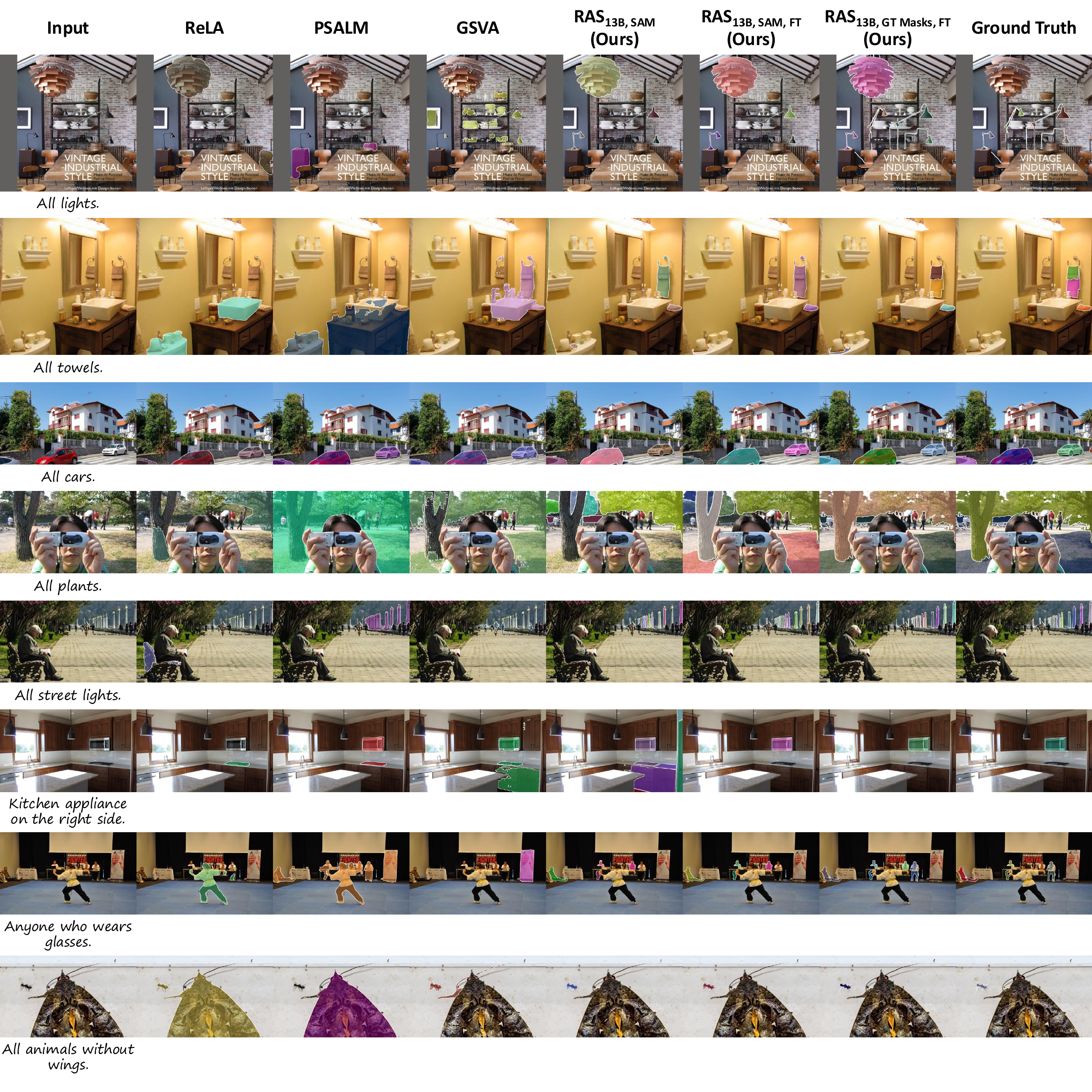}
    \caption{\textbf{Qualitative comparison on \ourgooddata.}}
    \label{fig:qual}
\end{figure*}

\noindent In Figure~\ref{fig:qual}, we provide additional visualized results of applying \ourmodel and other GRES models in the ORES task. \ourmodel (both before and after finetuned on \ourgooddata) achieves better results on \ourgooddata than all previous GRES models, which is consistent with our quantitative evaluation in Table~\ref{tab:ourgooddata} of the main paper.

\end{document}